\documentclass[10pt,twocolumn,letterpaper]{article}

\usepackage{cvpr}
\usepackage{times}
\usepackage{epsfig}
\usepackage{graphicx}
\usepackage{amsmath}
\usepackage{amssymb}
\usepackage{caption}
\usepackage{subcaption}
\usepackage{makecell}

\usepackage[breaklinks=true,bookmarks=false]{hyperref}

\cvprfinalcopy 

\def\cvprPaperID{10977} 

\begin{document}

\title{Texture Transform Attention for Realistic Image Inpainting}

\author{Yejin Kim\\
LG Electronics\\
{\tt\small yejin726.kim@lge.com}
\and
Manri Cheon\\
LG Electronics\\
{\tt\small junwoo.lee@lge.com}
\and
Junwoo Lee \\
LG Electronics\\
{\tt\small manri.cheon@lge.com}
}

\thispagestyle{empty}
\makeatletter
\let\@oldmaketitle\@maketitle
\renewcommand{\@maketitle}{\@oldmaketitle

\begin{center}
  \includegraphics[width=\linewidth] {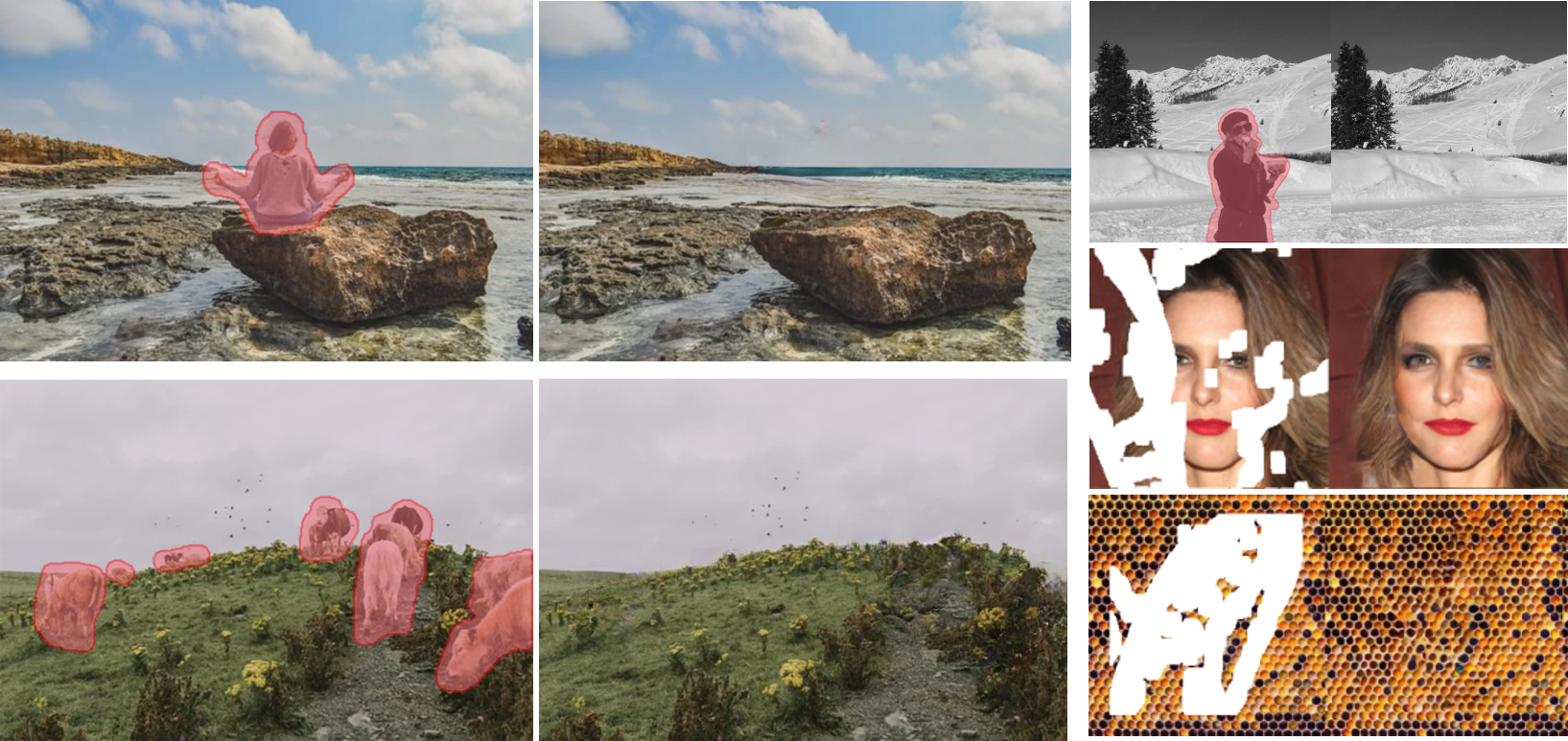}
   \captionof{figure}{Image inpainting results generated by the proposed system build on Texture Transfer Attention. Each triad displays the original image, the image with the damaged image masked in white, and the result of image inpainting. The results show high performance in expressing texture detail on variety of images. }
   
   \label{fig:fig01}
   \end{center}
   \bigskip}
\makeatother

\maketitle

\begin{abstract}
 Over the last few years, the performance of inpainting to fill missing regions has shown significant improvements by using deep neural networks. Most of inpainting work create a visually plausible structure and texture, however, due to them often generating a blurry result, final outcomes appear unrealistic and make feel heterogeneity. In order to solve this problem, the existing methods have used a patch based solution with deep neural network, however, these methods also cannot transfer the texture properly. Motivated by these observation, we propose a patch based method. Texture Transform Attention network(TTA-Net) that better produces the missing region inpainting with fine details. The task is a single refinement network and takes the form of U-Net architecture that transfers fine texture features of encoder to coarse semantic features of decoder through skip-connection. Texture Transform Attention is used to create a new reassembled texture map using fine textures and coarse semantics that can efficiently transfer texture information as a result. To stabilize training process, we use a VGG feature layer of ground truth and patch discriminator. We evaluate our model end-to-end with the publicly available datasets CelebA-HQ and Places2 and demonstrate that images of higher quality can be obtained to the existing state-of-the-art approaches
\end{abstract}

\section{Introduction}
 Image inpainting is an approach to plausibly synthesize alternative contents into missing regions \cite{ref04} damaged or non-critical information.  In computer vision, image inpainting has been focused on challenging topics and has produced a considerable progress over the decades \cite{ref01, ref03, ref08, ref16, ref19} and it has been applied to many tasks such as old photo restoration, image super resolution, crop/stitching, video inpainting and many others. 
 
 The core challenge of image inpainting is to generate high-level semantically-reasonable and visually-realistic texture details for the missing regions \cite{ref01, ref09, ref12, ref20, ref29, ref34}. Existing approaches are roughly divided into two broad groups: a texture synthesis approach using a low-level image feature and a feed-forward generative model using deep convolution networks. The texture synthesis approach \cite{ref03, ref06} can synthesize plausible stationary texture, via level-diffusion and patch-based algorithm. The patch-based algorithm \cite{ref03} iteratively searches a similar patch in background and pastes it into the missing region to synthesize a visually-realistic result. This approach works especially well simple composition and complex textures such as natural scene, however, it cannot hallucinate a novel image that contain the missing regions with high-semantic context or does not have adequate patches in the background. To solve this problem, the feed-forward generative model approach \cite{ref10, ref11, ref12, ref15, ref17, ref29, ref34} proposes to encode the semantic context of image into a feature space using deep neural network and decode the semantic patch to generate semantically-reasonable results. Unfortunately, however, this approach looks visually-blurry due to the loss of details caused by the repeated convolutions and poolings.
 
In order to ensure high-quality image inpainting performance, we propose a method of extracting the feature of each multi-layer, and efficiently transmitting them to decode the features back into the result. First we adopt U-Net structure \cite{ref26} that deliver multi feature from encoder to decoder via skip-connection. Second, we propose a Texture Transform Attention (TTA) module to efficiently transfer texture information to the result. The conventional patch-based networks \cite{ref29, ref31, ref35} method calculates the similarity of each patch through softmax in channel-wise and does summation by each weights. However, if there are many similar patches in the background, multiple patches are summated based on similar weights and the result appears blurry. To solve this problem, we search for the most similar patch and solely use the index and similarity weight on the patch to reflect in the result. Our method also applies a kernel of the same size to different resolutions for each layers, and changes the receptive field size to extract visual information such as semantic context of lower layers and texture details of upper layers. Third, we propose a highly simple approach of synthesis texture without using complex fusion or summation that cause blur. 

This network is one refinement network for fast and accurate training to learn high-level contexts for realistic texture synthesis. The main pipeline without the skip-connection generates the contextual structure in the missing regions and synthesizes the texture components transferred through the Texture Transform Attention for each layer. The TTA-Net is optimized by minimizing the reconstruction loss, adversarial GAN loss \cite{ref07}, VGG based perceptual loss \cite{ref13} and style loss \cite{ref30}. 

Experiments were conducted with publicly available datasets that could well represent inpainting performance such as faces, textures, and natural objects. Example results are shown in Figure \ref{fig:fig01}.
We highlight our contributions as follow:
\begin{itemize}
\item We propose a novel network utilizing a U-Net structure that directly transfers the encoded texture to the decoder by adding skip-connection to the already verified encode-decode image inpainting network. 
\item For the more efficient texture transfer, we propose the Texture Transform Attention module that searches the most similar patch. The TTA module find the index and similarity weight of the patch, reassembles the texture accordingly, and deliver it to the decoder. 
\item Our model namely, TTA-Net, can synthesize an image of more fine texture by iterative application of deep and shallow textures using a feature synthesis module. 
\item Our feed-forward generative network achieves high-quality inpainting result on variety of challenging datasets including CelebA faces \cite{ref36}, CelebA-HQ faces \cite{ref14}, DTD textures \cite{ref05} and Places2 \cite{ref21}.
\end{itemize}

\begin{figure*}
\begin{center}
\includegraphics[width=\linewidth] {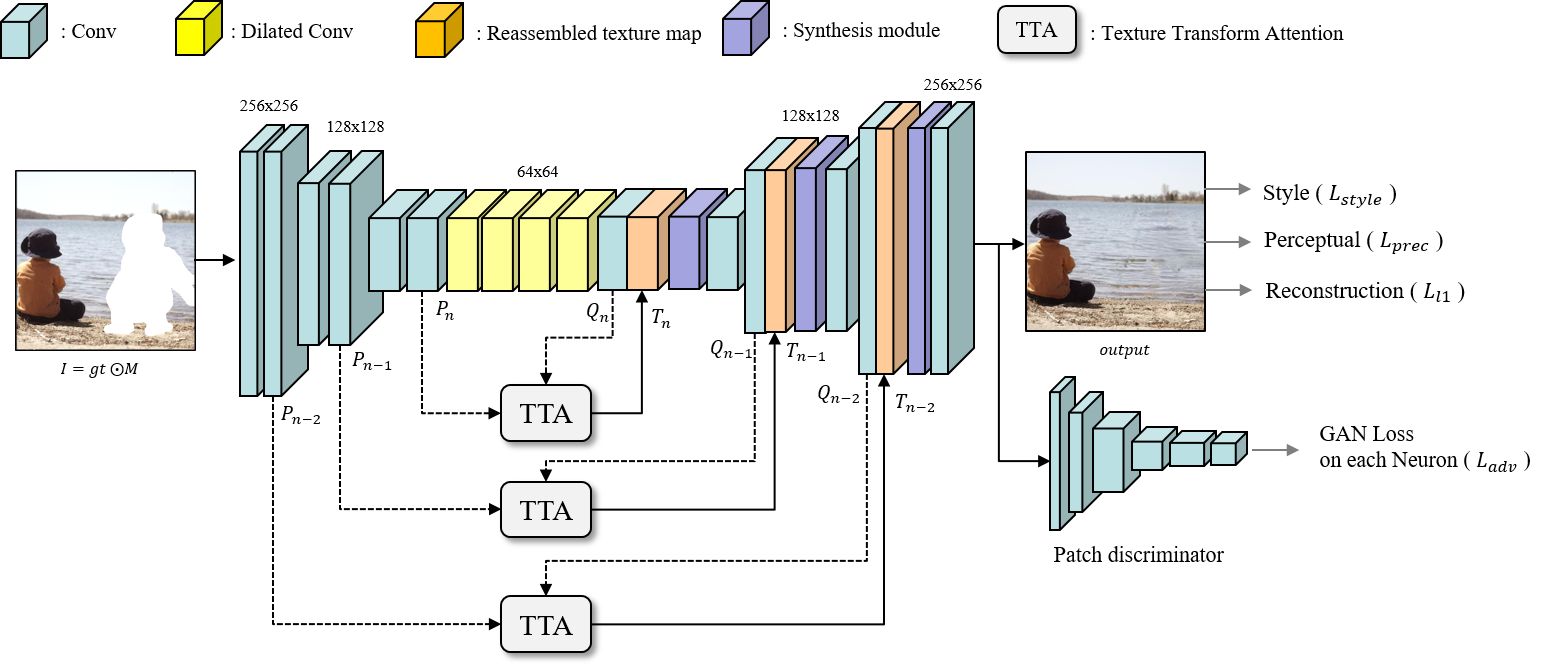}
\end{center}
   \caption{Overview of our framework with Texture Transform Attention module and feature synthesis module}
\label{fig:fig02}
\end{figure*}

\section{Related Work}
In computer vision, image inpainting has been focused on challenging topics and has produced significant progresses during the last decades \cite{ref01, ref03, ref08, ref17, ref19}. Inpainting researches can be largely divided into two categories: non-learning and learning inpainting approaches. The non-learning approach, is traditional diffusion-based or patch-based with low-level features. However, the learning approach is based on the deep neural network method that is most actively studied recently. This method learns convolution layer that predict the contents and pixels of the missing regions.

Traditional non-learning approaches such as \cite{ref01, ref04, ref05, ref06} can either propagate surrounding information or copying information from similar patches in the background to fill in missing regions. These methods are effective for stationary and repetitive texture information, but are limited only for non-stationary data that are locally unique. Huang et al. \cite{ref09} blended the known regions into target regions to minimize discontinuities. Simakov et al. \cite{ref18} proposed a bidirectional patch similarity-based scheme to better model anomalous visual data for re-targeting and inpainting applications. However, the approaches \cite{ref09, ref18} require very high expensive operation that dense computation of patch similarity. To solve this problem, Barnes et al. \cite{ref03} proposed a Patch-Match method, the fastest neighboring field algorithm using random initializations. Patch-Match shows a significantly better performance before the emergence of learning-based methods, and has been applied to a number of image editings.

In recent years, as researches on deep learning have been actively conducted, the paradigm of image inpainting has also been changed based on GAN-based approaches. The first deep neural networks for inpainting, Context encoder \cite{ref19}, firstly train deep neural networks for inpainting large holes, proposed a method of filling the missing regions with semantic information through feature learning and adversarial loss with novel encoder-decoder pipeline. However, it performs poorly in generating fine-detailed textures. Iizuka et al. \cite{ref10} proposed a generative model for high-resolution images using local and global discriminator and expanded the receptive field using dilated convolution.  However, additional post-processing step is required to maintain color consistency near hole boundaries. Yu et al \cite{ref29} proposed a generative network that create stacks that fill the pixels of missing regions with similar patches from the background to ensure color and texture consistency in the newly created areas and the surroundings. Pconv \cite{ref25} is designed to eliminate the mask effect through re-normalization by distinguishing the valid pixels of irregular masks. Yu et al. \cite{ref31} suggested learning of the dynamic feature selection mechanism for each channel and spatial location, which provides better visual results. PEN-net \cite{ref35} proposed a network that can effectively transmit high-semantic information from encoder to decoder by using a U-net structure that utilizes skip connection for multiple layers.

\begin{figure}[t]
\begin{center}
\includegraphics[width=\linewidth] {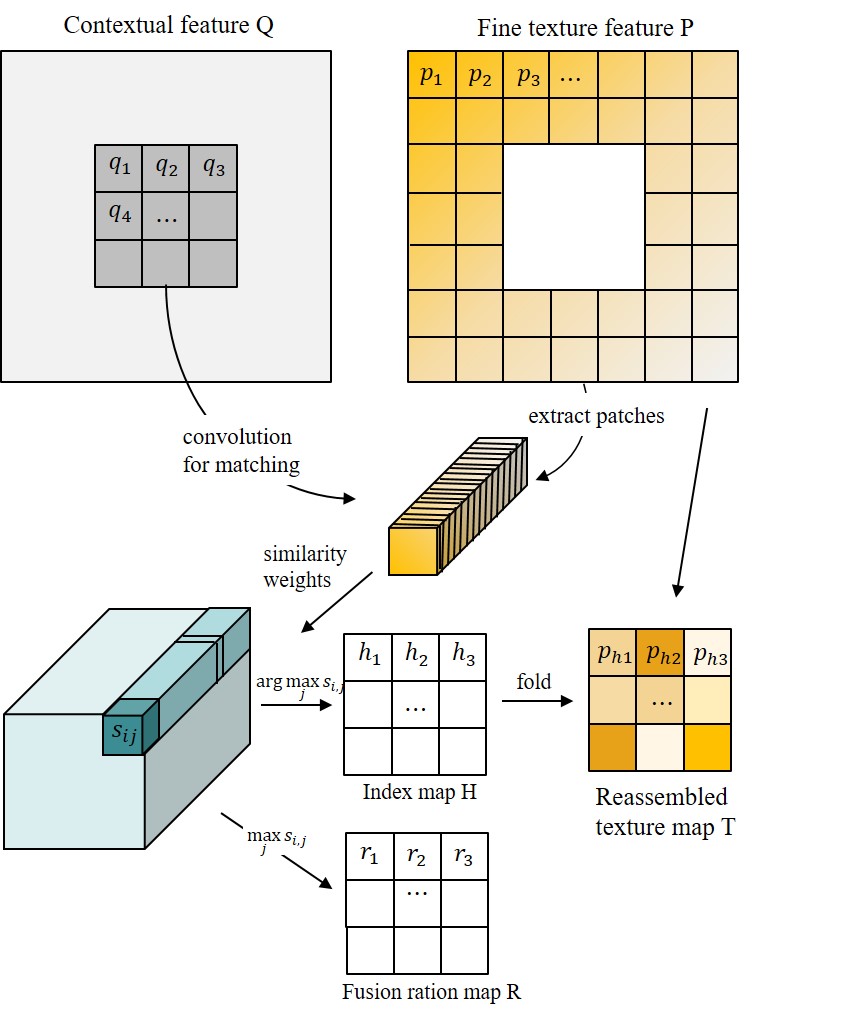}
\end{center}
   \caption{Illustration of the Texture Transform Attention layer. First, unfold the context feature and fine texture feature to same size for calculating the similarity weight (as convolution filters).  These similarity weight of all patches are compared channel-wise compared to find the index and weight of the most similar patch. Then we generate reassembled texture map by folding texture features according to the index map. The texture map and weight map are sent to the feature synthesis module and synthesized with the context feature.}
\label{fig:fig03}
\end{figure}

\section{Approach}
In this section, we introduce the proposed Texture Transformer Attention for Realistic Image. We first describe the overview of an inpainting network in Figure \ref{fig:fig02} and details of Texture Transformer Attention in Figure \ref{fig:fig03}.
\subsection{The Overall Pipeline}
The overall pipeline of proposed TTA-Net mechanism is illustrated in Figure \ref{fig:fig02}. This framework is a single refinement network that consist of with 4 modules, i.e., a encoder feature extractor, a Texture Transform Attention with skip-connection, a texture fusion multi decoder and a discriminator. The TTA-Net is a U-Net structure based on performance verified in-painting model \cite{ref11, ref19, ref25}, which can extract multi-layered latent features from encode and transmits them to decode through skip-connection. The two-stage network, which is widely used in recent years, is unstable in training due to its long parameter length. To improve training stability and maintain semantic performance, we propose therefore a single refinement network that applies VGG losses. The convolutions used in this framework(Figure \ref{fig:fig02}), uses gated convolutions \cite{ref31}, which are excellent for removal of mask effects, the discriminator and TTA module on the other hand use vanilla convolution.

 The feature extractor in encoder works at each resolution to extract a compact latent feature that is not damaged by iterative convolutions and poolings. As the compact latent features encode the multi feature information, high-level semantics of the context and low-level texture details are decoding via skip-connection. Dilated convolutions \cite{ref11, ref23} that are performed several times to fill a missing hole, creates a coarse semantic context feature that becomes the positional reference for generation of the Texture Transform Attention. The TTA module compares the generated semantic context feature and fine texture features transmitted from the encoder, reconstructs texture map using the index map of most similar patches. In decoder, a reassembled texture maps and context feature are synthesized multiple times to produce the result.
 
We adopt a patch discriminator \cite{ref31, ref37} for faster training and more stability. Also we use reconstruction loss and VGG loss \cite{ref13, ref30} that compare the ground truth and output of network.

\subsection{Texture Transform Attention}
The proposed Texture Transform Attention is to compensate for the blurry decoding context information by using the encoded information with various fine features. Conventional attention methods usually use the sum of similarity weights \cite{ref29, ref31, ref37}, but if there are many similar patches (ground, sand, bushes, etc.), similar weights will overlap, resulting in a blur. To overcome blur issue, we refer to a reference SR \cite{ref32, ref33, ref38} method of swap with or add fine texture components to the coarse image. When applied to inpainting network, we can assume that the coarse image is a context feature $Q$ created by dilated convolutions or upsampling, and fine texture is a undamaged feature $P$ extracted from encoder. Therefore, we aim to compare two different features as shown in Figure \ref{fig:fig03}, reassemble the texture feature $P$ semantically similar and synthesize it to the context feature $Q$. 

{\bf Relevance Embedding}.  Relevance embedding aims to embed the relevance between the encoded texture features $P$ and decoded context features $Q$  by calculating the matching score. We unfold $P$ (without missing regions) and $Q$(missing regions) into 5x5 size patches of $p_j \in P$ and $q_i \in Q$ respectively. By unfolding different resolutions into patches of same size and changing the receptive fields of each layer, different attention maps are created to help consist inpainting performance even when a new semantic context is created. When calculating the similarity between $P$ and $Q$, we cut the resolution in half and reduce the patch size to 3x3 to speed up the comparison and save resources. We calculate the relevance $s_{i,j}$ between these two patches by normalized inner product (cosine similarity):
\begin{equation}
s_{i,j}=\left<\frac{p_i}{\| p_i \|}, \frac{q_j}{\| q_j \|}\right>
\end{equation}
where $s_{i,j}$ represents similarity of patch of fine texture components $p_i$ and coarse semantic patch $q_j$. This similarity $s_{i,j}$ is used to create a reassembled texture map $T$ and fusion of $Q$ and $T$.

{\bf Feature Swapping}.  We propose a method of creating reassembled texture map $T$ that reconstruct the details of the fine texture $P$ in the form of the semantic feature $Q$. The conventional attention methods take a weighted sum of channel-wise patches, but this weakens the transmission performance of the fine texture and lacks detail. Synthesizing solely the most relevant patch $p_j$ at the position of $q_i$ shows better performance in detail expression. 

First, channel-wise comparison of the similarity $s_{i,j}$ is performed on each patch to find the $p_j$ patch most similar to $q_i$ and generate an index map $H$ for location.
\begin{equation}
h_i=\text{arg}\max_j s_{i,j}
\end{equation}
where $h_i$ is the index representing the location of the most relevant patch in the fine texture $P$ for the $i$-th element patch of the semantic feature $Q$. By applying the index map $H$ to transfer the patches of fine texture $P$, we can generate a reassembled texture layer $T$ that can be applied to decoding. 
\begin{equation}
t_i=p_{h_i}
\end{equation}
where $t_i$ denotes the value of $T$ in the $i$-th position, which is selected from $h_i$-th position of $P$. 

 Summary, reassembled texture map $T$ is obtained by reconstruction texture feature $P$ according to the semantic shape of $Q$, and transmitted to the decoder and reflect in the result

\begin{figure*}
\begin{center}
\begin{subfigure}[]{0.18\textwidth}
\includegraphics[width=\textwidth] {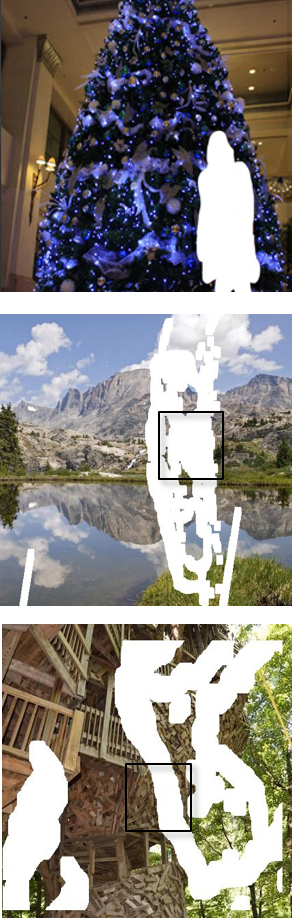}
\caption{Input}
\label{fig:places2_input}
\end{subfigure}
\hfill
\begin{subfigure}[]{0.19\textwidth}
\includegraphics[width=\textwidth] {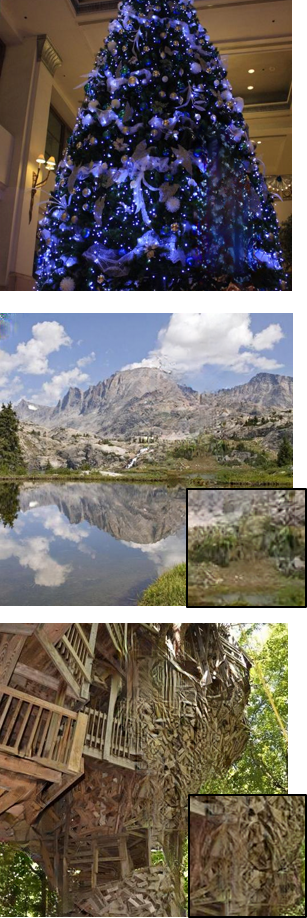}
\caption{PConv(online demo)}
\label{fig:places2_pconv}
\end{subfigure}
\hfill
\begin{subfigure}[]{0.191\textwidth}
\includegraphics[width=\textwidth] {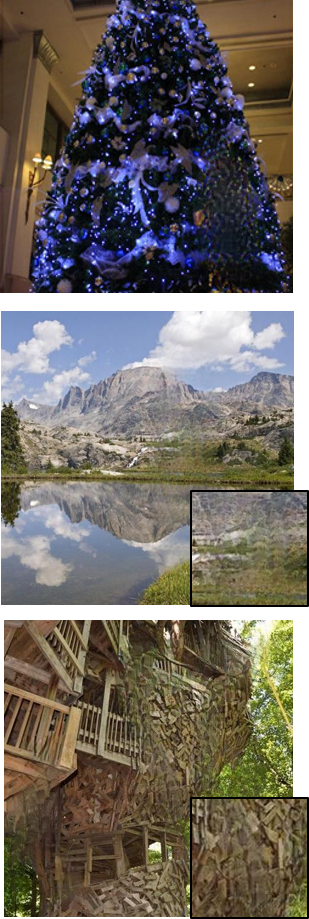}
\caption{EdgeConnect)}
\label{fig:places2_EC}
\end{subfigure}
\hfill
\begin{subfigure}[]{0.194\textwidth}
\includegraphics[width=\textwidth] {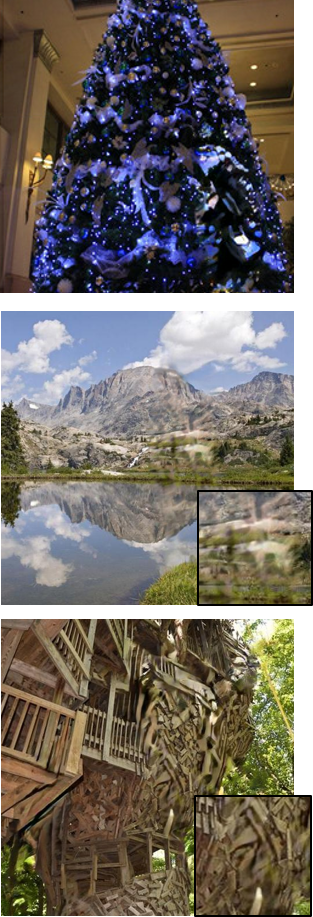}
\caption{DeepFillv2}
\label{fig:places2_DF2}
\end{subfigure}
\hfill
\begin{subfigure}[]{0.193\textwidth}
\includegraphics[width=\textwidth] {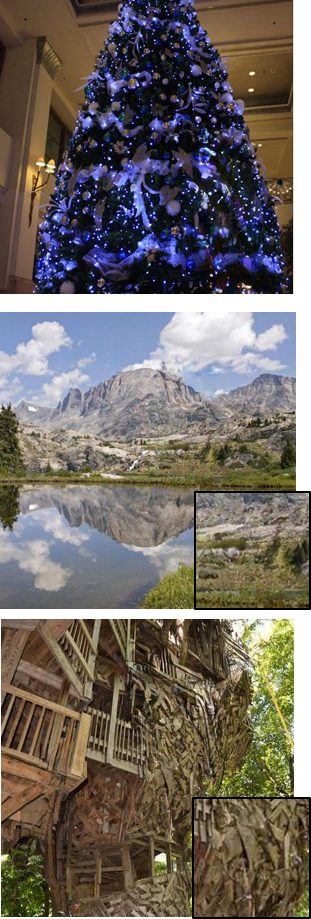}
\caption{Ours}
\label{fig:places2_Ours}
\end{subfigure}
\end{center}
   \caption{Comparison of qualitative results with models that proven performance on Places validation sets. Best viewed (especially texture) with zoom-in.(a) Input is ground truth with mask. (b) Partial convolution \cite{ref25}. (c) EdgeConnect \cite{ref34}. (d) Gated conv \cite{ref31}. (f) Ours.}
\label{fig:fig04}
\end{figure*} 
 
\subsection{Similarity weight texture synthesis}
{\bf Fusion ration map}.  We propose a method to synthesize a semantic feature $Q$ and a reassembled texture $T$ in the decoder. The context features that created in decoder may not have similar patches in the background or have low similarity, so we need to ensure that not all patches are synthesized at the same ratio for preventing a ghost effect. To solve the  problem, our method  refers to the similarity weight value $s_{i,j}$ calculated above and does a channel-wise comparison to search the highest ratio $r_i$ for fusion. The ratio map $R$ represent confidence of the reassembled texture map for each position in $T$ :
\begin{equation}
r_i=\max_j s_{i,j}
\end{equation}
where $r_i$ denotes the $i$-th position of fusion ratio map of $R$.

{\bf Similarity Fusion}. The fusion ratio map $R$ is used to synthesize the decoder context feature $Q$ and reassembled texture map $T$. Instead of directly applying $R$ to $T$, we first concatenate $Q$ and $T$ and perform convolution layer.  The fused features are element-wisely multiplied by the ratio map $R$ and added back to $Q$. This operation can be represented as:
\begin{equation}
F_{fus}=F+Conv(Concat(F,T))\odot R
\end{equation}
where $F$ is same to the semantic feature $Q$ and $F_{fus}$ indicated the synthesized output features. $Conv$ represents a convolution layer and $Concat$ indicates a concatenation operation. The operator $\odot$ denotes an element-wise multiplication between feature maps. 

However, since each patch is simply multiplied by a different ratio $r_i$ and added to $F$, a color distortion occurs that results checkerboard effect as shown in Figure \ref{fig:fig07}.  To solve this problem, we used normalization factor $(1+R)$ and multiplied $F_{fus}$ by element-wise. 
\begin{equation}
F_{out}=F_{fus} \odot (1+H)^{-1}
\end{equation}
where $F_{out}$ indicates the synthesized output features. 

In summary, the Texture Transform Attention effectively transfers relevant fine texture $P$ to semantic context feature $Q$, making resulting images more realistic. 

\subsection{Training objective}
The factors considered when choosing loss are: 1) to improve the spatial structure of inpainted missing region, 2) to create a plausible visual quality of resulting image, and 3) to take advantage of rich texture from the encode feature extractor. Our objective function combines reconstruction loss  $\mathcal{L}_{rec}$, perceptual loss $\mathcal{L}_{per}$, adversarial loss $\mathcal{L}_{adv}$ and style loss $\mathcal{L}_{style}$. We experimentally tested the appropriate hyperparameters $\lambda$ and experiments with the conditions of $\lambda_{rec}=\lambda_{adv}=1$, $\lambda_{per}=0.1$ and $\lambda_{style}=100$.  
\begin{equation}
\begin{split}
\mathcal{L}_{overall}=\lambda_{rec}\mathcal{L}_{rec}+\lambda_{adv}\mathcal{L}_{adv}\\
+\lambda_{per}\mathcal{L}_{per}+\lambda_{style}\mathcal{L}_{style}
\end{split}
\end{equation}

{\bf Reconstruction loss}. The reconstruction loss contributes to create an approximate shape by comparing the predicted image to ground truth. We can generate a more accurate $I_{pred}$ by adopting an $l_1$-norm instead of a MSE
\begin{equation}
\mathcal{L}_{rec}=\|I_{gt}-I_{pred}\|
\end{equation}

{\bf Adversarial loss}. Adversarial loss can significantly improve the structural/visual quality of this synthesized image. We adopt the SN-PatchGAN loss \cite{ref31} for more stable training and improve semantic information and local texture. $D^{sn}$ denotes a spectral-normalized discriminator and $G$ indicates an inpainting network that receives an incomplete image $z=I_{gt}\odot (1-M)$ as input.
\begin{equation}
\mathcal{L}_G=-\mathbb{E}_{z\sim(z)}[D_{sn}(G(z))]
\end{equation}
\begin{equation}
\begin{split}
\mathcal{L}_D=\mathbb{E}_{x\sim\mathbb{P}_{data}(z)}[ReLU(1-D^{sn}(x)]\\
+\mathbb{E}_{z\sim\mathbb{P}_z(z)}[ReLU(1+D^sn(G(z))]
\end{split}
\end{equation}

{\bf Perceptual loss}. We only use a single segmentation network for precise training, rather than a two-stage network consisting of coarse and refinement network that are widely used. We adopted perceptual loss \cite{ref13} for visual quality improvement. $\mathcal{L}_{per}$ defines the Euclidean distance 

value of activation map using pre-trained VGG16, penalizing if the label and perceptual similarity are not close. 
\begin{equation}
\mathcal{L}_{per}=\frac{1}{V}\sum_{t=1}^C\|\phi_i^{vgg}(I_{gt})-\phi_i^{vgg}(I_{pred})\|_1
\end{equation}
where $V$ and $C$ indicate the volume and channel number of the feature map, respectively, and $\phi_i^{vgg}$ is the activation map of the $i$-th layer of the pre-trained VGG network. 

{\bf Style loss}. We adopted style loss \cite{ref30} to create more realistic texture detail. Style loss computes the difference between covariance in the activation map and uses a gram matrix $G$ to avoid checkerboard artifacts due to convolution layers. 
\begin{equation}
\mathcal{L}_{style}=\|G(\phi_i^{vgg}(I_{comp})-G(\phi_i^{vgg}(I_{pred}))\|_1
\end{equation}

\subsection{Segmentation mask generation}
Inpainting is often used as an object removal, and we can therefore assume that the object to be person or something user do not want. Therefore, to create a mask that real user will make, we used an instance segmentation \cite{ref41} algorithm to segment the objects in the places2 dataset and create a mask.  Specially, the missing region of the mask is  $10\sim 40\%$ of the total size, and the category is limited to people, animal, statues and etc. About 1,000 masks were created and used as a validation mask set. In the training phase, we conducted various case studies using NVIDIA random masks \cite{ref25}, and we aimed to improve user-friendly performance using the created segmentation mask for validation.

\begin{table}[t]
\begin{center}
\begin{tabular}{c  c  c  c  c}
\Xhline{3\arrayrulewidth}
Method & L1 loss† & MS-SSIM¶ & FID† & LPIPS† \\
\hline
\hline
DeepFillv1 & 6.99 & 80.26 & 110.9 & 0.1195 \\
DeepFillv2 & 4.46 & 85.58 & 48.62 & 0.0687\\
EdgeConnect & {\bf 3.8} & {\bf 86.72} & 44.96 & 0.0609\\
Ours & 4.36 & 85.3 & {\bf 43.52} & {\bf 0.0595}\\
\Xhline{3\arrayrulewidth}
\end{tabular}
\caption{Quantitative comparison on validation images of Places2 with L1 Loss, MS-SSIM, FID and LPIPS. We use free-form mask where the missing region is 10 to 40\% of mask size. 
† Lower is better. ¶ Higher is better. }
\label{tab:tab01}
\end{center}
\end{table}

\section{Experimental Results}
The proposed system was evaluated of Places2 \cite{ref21}, Celeb-HQ faces \cite{ref15} and DTD textures \cite{ref05}, in term of quantitative and qualitative. Our model was trained using an NVIDIA RTX GPU with a 256x256 resolution size with batch size of 16. In the training phase, the datasets of CelebA-HQ and DTD were downsampled to 256x256 resolution from their original size and Places2 is randomly cropped for texture learning. Basically the model was trained with pytorch v3.6, CUDNN v7.0, CUDA v10.0. It took about 0.15 seconds to test a 512x512 resolution image, regardless of a hole size, and no additional pre- or post-processing is required.

\begin{figure}[t]
\begin{center}
\begin{subfigure}[]{0.152\textwidth}
\includegraphics[width=\textwidth] {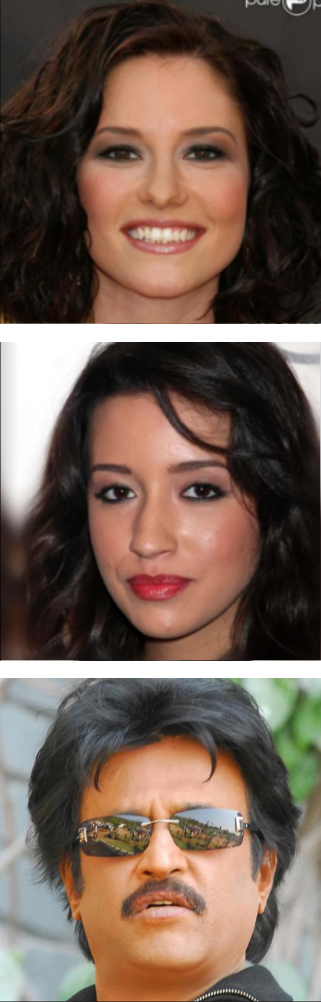}
\caption{Ground truth}
\label{fig:face_gt}
\end{subfigure}
\hfill
\begin{subfigure}[]{0.152\textwidth}
\includegraphics[width=\textwidth, height=3.12\textwidth] {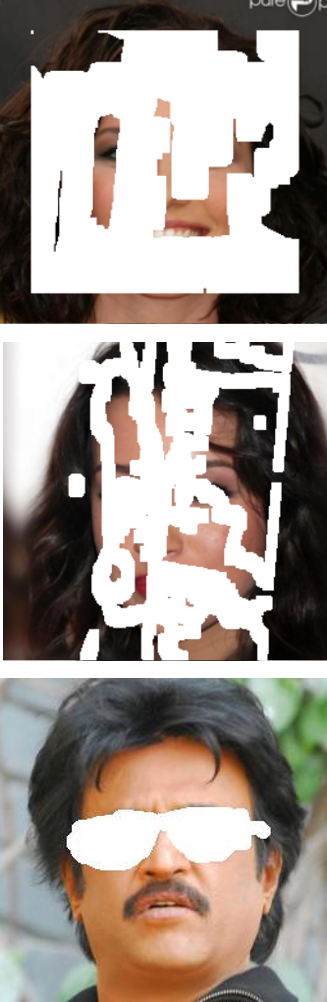}
\caption{Input}
\label{fig:face_input}
\end{subfigure}
\hfill
\begin{subfigure}[]{0.152\textwidth}
\includegraphics[width=\textwidth, height=3.12\textwidth] {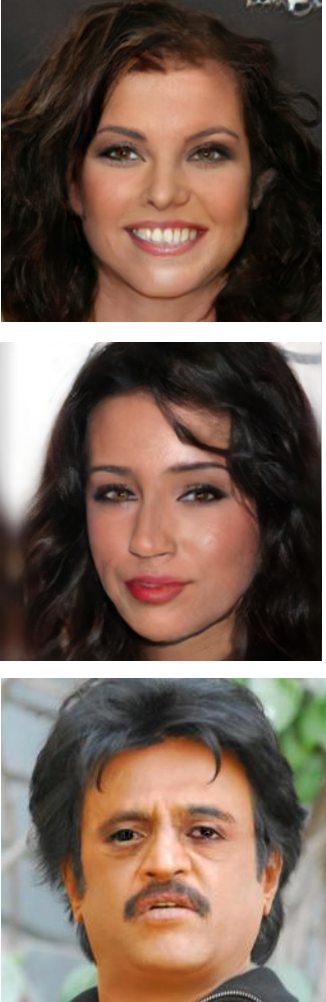}
\caption{Ours}
\label{fig:face_ours}
\end{subfigure}
\end{center}
   \caption{Example results generated by the proposed network on CELEB-HQ.}
\label{fig:fig05}
\end{figure} 

\subsection{Quantitative Results}
Like other image generation tasks, image inpainting lacks good quantitative evaluation metrics. Even if the inpainted region is not the same as the ground truth, it is acceptable to the user if it has a plausible structure and texture. The results listed in Table \ref{tab:tab01} show the performance of our model and baseline on 512x512 resolution validation image of Places2 with free-form random masks. 

We report our evaluation in terms of reconstruction $\mathcal{L}_1$ loss, multi-scale structural similarity(MS-SSIM) \cite{ref28}, Frechet Inception Distance(FID) \cite{ref27} and Learned Perceptual Image Patch Similarity(LPIPS) \cite{ref39}. The $\mathcal{L}_1$ roughly reflect model’s performance to reconstruct the original image content. MS-SSIM extracts and evaluates the similarity of structural information from paired images at multiple scales, providing results that approximate human visual perception. These metrics assume pixel-wise independence, which may assign favorable scores to perceptually inaccurate results. Therefore, we include evaluation metrics that use a deep feature based on human perception. FID measures a Wasserstein-2 distance between the feature space representations of real and inpainted images using a pre-trained Inception-V3 model \cite{ref40}. LPIPS uses pre-trained VGG to measure the perceptual similarity distances to human judgment.

As shown in Table \ref{tab:tab01}, our proposed method shows that $\mathcal{L}_1$ loss and MS-SSIM are similar to other models and better in FID and LPIPS. These mean that our model have high-semantic inpainting performance similar to other models, but have better texture expression.

\begin{figure}[t]
\begin{center}
\includegraphics[width=\linewidth] {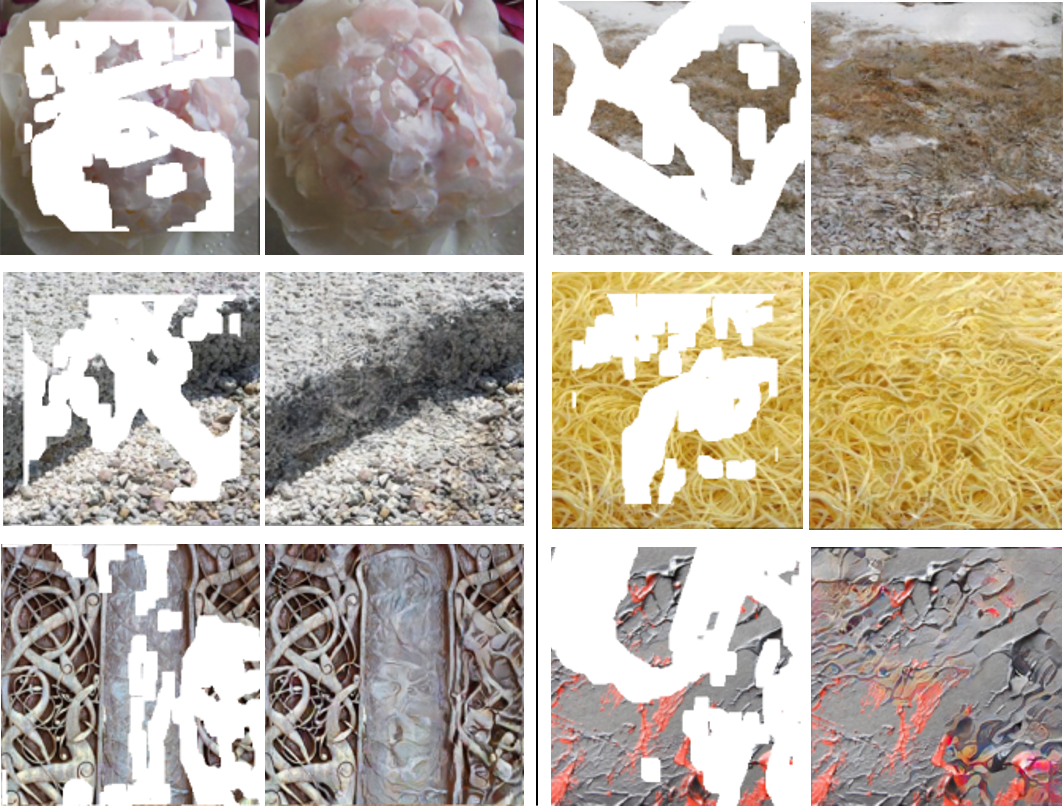}
\end{center}
   \caption{Example results generated by the proposed network on DTD Texture. (Left is input and right is result)}
\label{fig:fig06}
\end{figure} 

\begin{figure*}
\begin{center}
\begin{subfigure}[]{0.18\textwidth}
\includegraphics[width=\textwidth] {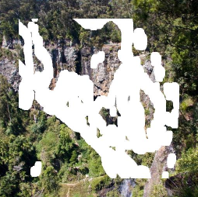}
\caption{Input}
\label{fig:NF1}
\end{subfigure}
\hfill
\begin{subfigure}[]{0.4\textwidth}
\includegraphics[width=\textwidth] {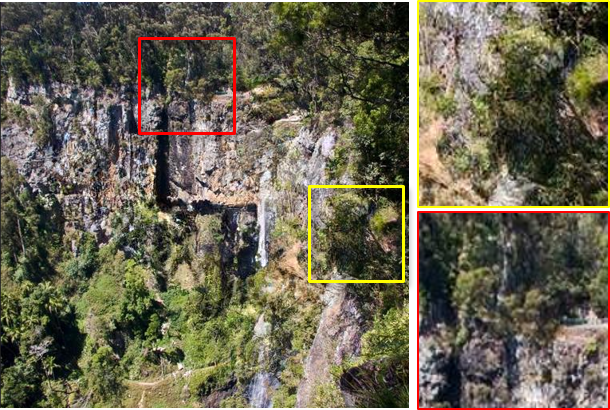}
\caption{With Normalization factor}
\label{fig:NF2}
\end{subfigure}
\hfill
\begin{subfigure}[]{0.41\textwidth}
\includegraphics[width=\textwidth] {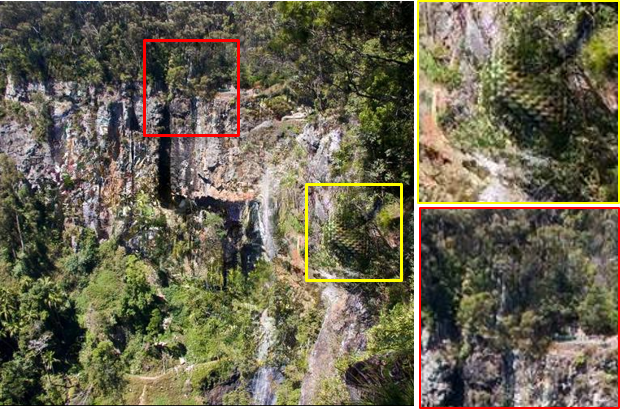}
\caption{Without Normalization factor}
\label{fig:NF3}
\end{subfigure}
\end{center}
   \caption{Example of results that normalize factor comparison}
\label{fig:fig07}
\end{figure*}

\subsection{Qualitative Results}
We compared the proposed model with previous state-of-the-art approaches \cite{ref25, ref29, ref31, ref34}. In order to compare the texture generating performance, a qualitative comparison was performed by challenging images from the places2 dataset with complex and irregular textures with an original size. In Figure \ref{fig:fig04} the entire image and enlarged image patch are displayed together to compare the semantic context and texture of the image. Figure \ref{fig:fig05} and Figure \ref{fig:fig06} show the results of our model using CelebA datasets and DTD texture datasets. We use user custom masks and free-form random masks to compare various cases.

As shown in Figure \ref{fig:fig04}: Comparison of qualitative results with models that proven performance on Places validation sets. Best viewed (especially texture) with zoom-in.(a) Input is ground truth with mask. (b) Partial convolution \cite{ref25}. (c) EdgeConnect \cite{ref34}. (d) Gated conv \cite{ref31}. (f) Ours., all models have succeeded in making the missing regions plausible, but they have different results. The result of EC \cite{ref34} was used with similar texture patches repeatedly around the missing region without taking into account the semantic context, giving it a heterogeneous feel. The result of GC \cite{ref31} has a semantic structure, but the texture looks blurry compared to the surroundings as it accumulates several similar patches. On the other hand, the result of our model has a composition that considers the semantic context of surrounding area, and it can be confirmed that a photo-realistic result appears by creating a texture similar to a background property. Our model is trained at a small resolution of 256x256 like other comparison models, but it shows better results even with test image in a larger size than the training size, as the surrounding features are transferred the low-level layer to the high-level layer in the network. 

Also, as shown in Figure \ref{fig:fig06} and Figure \ref{fig:fig05}, our model works well with human faces or complex and repetitive textures.

\subsection{Ablation Study}
{\bf Synthesis module comparison}. We investigated the effectiveness of the synthesis module by comparing to other synthesis methods including a cross-scale feature synthesis \cite{ref32}, commonly used concatenation method \cite{ref29, ref31} and dilated convolution \cite{ref35}. The concatenation method is difficult to predict or control how the texture will be reflected in the results simply by attaching texture information to the existing layer. The dilated convolution and cross-scale feature synthesis method naturally combine multiple textures and produce visually sharp results. However, we referenced a soft-attention approach \cite{ref32} that uses a similarity weight because the performance improvement obtained in these modules is not significant compared to the consumed resources.

{\bf Whether to use normalization factor}. We added normalized factor to our synthesis module. If we simply multiply the similarity weight pixel-wise, the result will have different color level for each region, generating in checkerboard artifacts. As we apply the normalize factor, color normalization can be performed according to the fusion ratio of each patch, which can achieve natural results. In Figure \ref{fig:fig07}, we can see the comparison of results according to use of the normalization factor. 

\section{Conclusion}
In this paper, we proposed a novel image inpainting system based on an end-to-end U-Net generative network with a Texture Transform Attention which efficiently transfers a fine texture from the encoder to the decoder. We showed that the TTA module significantly improves the texture representation while preserving the semantic context of result. Also, we proposed a highly simple and effective sysnthesis module to reflect fine texture in the results. Quantitative results and, qualitative comparisons demonstrated the superiority of our proposed TTA-Net. As a future work, we plan to improve the proposed network to image with a higher resolution and modify it to work well with videos. 


{\small
\bibliographystyle{ieee_fullname}
\bibliography{egbib}

\begin{thebibliography}{10}\itemsep=-1pt

\bibitem{ref03}
Connelly Barnes, Eli Shechtman, Adam Finkelstein, and Dan~B Goldman.
\newblock Patchmatch: A randomized correspondence algorithm for structural
  image editing.
\newblock {\em ACM Transactions on Graphics (TOG) (Proceedings of SIGGRAPH
  2009)}, 2009.

\bibitem{ref04}
M. Bertalmio, G. Sapiro, V. Caselles, and C. Ballester.
\newblock Image inpainting.
\newblock {\em Proceedings of the 27th annual conference on Com-puter graphics
  and interactive techniques}, page 417–424, 2000.

\bibitem{ref05}
M. Cimpoi, S. Maji, I. Kokkinos, S. Mohamed, and A. Vedaldi.
\newblock Describing textures in the wild.
\newblock {\em Proceedings of the IEEE Conference on Computer Vi-sion and
  Pattern Recognition}, page 3606–3614, 2014.

\bibitem{ref01}
Antonio Criminisi, Patrick P´erez, and Kentaro Toyama.
\newblock Region filling and object removal by exemplar-based im-age
  inpainting.
\newblock {\em TIP}, 13(9):1200–1212, 2004.

\bibitem{ref06}
A.~A. Efros and W.~T. Freeman.
\newblock Image quilting for texture synthesis and transfer.
\newblock {\em Proceedings of the 28th annual conference on Com-puter graphics
  and interactive techniques}, page 341–346, 2001.

\bibitem{ref07}
A.~A. Efros and T.~K. Leung.
\newblock Texture synthesis by nonparametric sampling.
\newblock {\em Computer Vision, 1999. The Proceedings of the Seventh IEEE
  International Conference}, 2:1033–1038, 1999.

\bibitem{ref30}
L.~A. Gatys, A.~S. Ecker, and M. Bethge.
\newblock Image style transfer using convolutional neural networks.
\newblock {\em Proceedings of the IEEE Conference on Computer Vision and
  Pattern Recognition (CVPR)}, page 2414–2423, 2016.

\bibitem{ref08}
I. Goodfellow, J. Pouget-Abadie, M. Mirza, B. Xu, D.Warde-Farley, A.~Courville
  S.~Ozair, and Y. Bengio.
\newblock Generative adversarial nets.
\newblock {\em Advances in neural information processing systems}, page
  2672–2680, 2014.

\bibitem{ref09}
J. Hays and A.~A. Efros.
\newblock Scene completion using millions of photographs.
\newblock {\em ACM Transactions on Graphics (TOG)}, 2007.

\bibitem{ref27}
Martin Heusel and Hubert Ramsauer.
\newblock Thomas unterthiner, bernhard nessler, and sepp hochreiter. gans
  trained by a two time-scale update rule converge to a local nash equilibrium.
\newblock {\em NeurIPS}, page 6626–6637, 2017.

\bibitem{ref41}
R. Hu, P. Dollar, K. He, T. Darrell1, and R. Girshick.
\newblock Learning to segment every thing.
\newblock {\em CVPR}, pages 4233--4241, 2018.

\bibitem{ref10}
J.-B. Huang, S.~B. Kang, N. Ahuja, and J. Kopf.
\newblock Image completion using planar structure guidance.
\newblock {\em ACM Transactions on Graphics (TOG)}, 33(4):129, 2014.

\bibitem{ref11}
S. Iizuka, E. Simo-Serra, and H. Ishikawa.
\newblock Globally and locally consistent image completion.
\newblock {\em ACM Transactions on Graphics (TOG}, 36(4):107, 2017.

\bibitem{ref37}
P. Isola, J. Zhu, T. Zhou, and A.~A Efros.
\newblock Image-to-image translation with conditional adversarial networks.
\newblock {\em Proceedings of the IEEE conference on computer vision and
  pattern recognition}, page 1125–1134, 2018.

\bibitem{ref12}
M. Jaderberg, K. Simonyan, A. Zisserman, and et al.
\newblock Spatial transformer networks.
\newblock {\em Advances in Neural Information Processing Systems}, page
  2017–2025, 2015.

\bibitem{ref13}
Y. Jeon and J. Kim.
\newblock Spatial transformer networks.
\newblock {\em Active convolution: Learning the shape of convolution for image
  classification}, {\em arXiv preprint arXiv:1703.09076}, 2017.

\bibitem{ref14}
J. Johnson, A. Alahi, and L. Fei-Fei.
\newblock Perceptual losses for real-time style transfer and su-per-resolution.
\newblock {\em European Conference on Computer Vision}, page 694–711, 2016.

\bibitem{ref15}
T. Karras, T. Aila, S. Laine, and J. Lehtinen.
\newblock Progressive growing of gans for improved quality, stabil-ity, and
  variation.
\newblock {\em arXiv preprint arXiv:1710.10196}, 2017.

\bibitem{ref16}
R. K¨ohler, C. Schuler, B. Sch¨olkopf, and S. Harmeling.
\newblock Mask-specific inpainting with deep neural networks.
\newblock {\em German Conference on Pattern Recognition}, page 523–534, 2014.

\bibitem{ref17}
A. Levin, A. Zomet, S. Peleg, and Y.Weiss.
\newblock Seamless image stitching in the gradient domain.
\newblock {\em Computer Vision-ECCV 2004}, page 377–389, 2004.

\bibitem{ref25}
Guilin Liu, Fitsum~A Reda, Kevin~J Shih, Ting-Chun Wang, Andrew Tao, and Bryan
  Catanzaro.
\newblock Image inpainting for irregular holes using partial convolu-tions.
\newblock {\em ECCV}, page 85–100, 2018.

\bibitem{ref36}
Z. Liu, P. Luo, X. Wang, and X. Tang.
\newblock Deep learning face attributes in the wild.
\newblock {\em Proceedings of International Conference on Computer Vision
  (ICCV)}, 2015.

\bibitem{ref34}
Kamyar Nazeri, Eric Ng, Tony Joseph, Faisal Qureshi, and Mehran Ebrahimi.
\newblock Edgeconnect: Generative image inpainting with adversari-al edge
  learning.
\newblock {\em arXiv preprint arXiv:1901.00212}, 2019.

\bibitem{ref18}
E. Park, J. Yang, E. Yumer, D. Ceylan, and A.~C. Berg.
\newblock Transformation-grounded image generation network for novel 3d view
  synthesis.
\newblock {\em arXiv preprint arXiv:1703.02921}, 2017.

\bibitem{ref19}
D. Pathak, P. Krahenbuhl, J. Donahue, T. Darrell, and A.~A. Efros.
\newblock Context encoders: Feature learning by inpainting.
\newblock {\em Proceedings of the IEEE Conference on Computer Vi-sion and
  Pattern Recognition}, page 2536–2544, 2016.

\bibitem{ref26}
Olaf Ronneberger, Philipp Fischer, and Thomas Brox.
\newblock U-net: Convolutional networks for biomedical image segmentation.
\newblock {\em MICCAI}, page 234–241, 2015.

\bibitem{ref20}
D. Simakov, Y. Caspi, E. Shechtman, and M. Irani.
\newblock Summarizing visual data using bidirectional similarity.
\newblock {\em Computer and Pattern Recognition, 2008. CVPR 2008. IEEE
  Conference on}, pages 1--8, 2008.

\bibitem{ref40}
C. Szegedy, V. Vanhoucke, S. Ioffe, J. Shlens, and Z. Wojna.
\newblock Rethinking the inception architecture for computer vision.
\newblock {\em Proceedings of the IEEE Conference on Computer Vision and
  Pattern Recognition (CVPR)}, page 2818–2826, 2016.

\bibitem{ref28}
Zhou Wang, Eero~P Simoncelli, and Alan~C Bovik.
\newblock Multiscale structural similarity for image quality assess-ment.
\newblock {\em ACSSC}, 2:1398–1402, 2003.

\bibitem{ref21}
C. Yang, X. Lu, Z. Lin, E. Shechtman, O. Wang, and H. Li.
\newblock High-resolution image inpainting using multi-scale neural patch
  synthesis.
\newblock {\em arXiv preprint arXiv:1611.09969}, 2016.

\bibitem{ref32}
Fuzhi Yang, Huan Yang, Jianlong Fu, Hontao Lu, and Bain ing Guo.
\newblock Learning texture transformer network for image su-per-resolution.
\newblock {\em Proceedings of the IEEE Conference on Computer Vision and
  Pattern Recognition (CVPR)}, pages 5791--5798, 2019.

\bibitem{ref23}
F. Yu and V. Koltun.
\newblock Multi-scale context aggregation by dilated convolutions.
\newblock {\em arXiv preprint arXiv:1511.07122}, 2015.

\bibitem{ref29}
Jiahui Yu, Zhe Lin, Jimei Yang, Xiaohui Shen, Xin Lu, and Thomas~S Huang.
\newblock Generative image inpainting with contextual attention.
\newblock {\em CVPR}, page 5505–5514, 2018.

\bibitem{ref31}
Jiahui Yu, Zhe Lin, Jimei Yang, Xiaohui Shen, Xin Lu, and Thomas~S Huang.
\newblock Image style transfer using convolutional neural networks.
\newblock {\em Free-form image inpainting with gated convolution}, {\em arXiv
  preprint arXiv:1806.03589}, 2018.

\bibitem{ref35}
Yanhong Zeng, Jianlong Fu, Hongyang Chao, and Baining Guo.
\newblock Learning pyramid-context encoder network for high-quality image
  inpainting.
\newblock {\em arXiv preprint arXiv:1904.07475}, 2019.

\bibitem{ref39}
R. Zhang, P. Isola, A.~A. Efros, E. Shecht-man, and O. Wang.
\newblock The unreasonable effectiveness of deep features as a perceptual
  metric.
\newblock {\em Proceedings of the IEEE Conference on Computer Vision and
  Pattern Recognition}, 2018.

\bibitem{ref33}
Zhifei Zhang, ZhaowenWang, Zhe Lin, and Hairong Qi.
\newblock Image super-resolution by neural texture transfer.
\newblock {\em CVPR)}, page 7982–7991, 2019.

\bibitem{ref38}
H. Zheng, M. Ji, H.Wang, Y. Liu, and L. Fang.
\newblock Cross-net: An end-to-end reference-based super resolu-tion network
  using cross-scale warping.
\newblock {\em European Conference on Computer Vision (ECCV)}, 2018.

\end{thebibliography}


\begin{thebibliography}{1}\itemsep=-1pt

\bibitem{Alpher02}
FirstName Alpher.
\newblock Frobnication.
\newblock {\em Journal of Foo}, 12(1):234--778, 2002.

\bibitem{Alpher03}
FirstName Alpher and FirstName Fotheringham-Smythe.
\newblock Frobnication revisited.
\newblock {\em Journal of Foo}, 13(1):234--778, 2003.

\bibitem{Alpher04}
FirstName Alpher, FirstName Fotheringham-Smythe, and FirstName Gamow.
\newblock Can a machine frobnicate?
\newblock {\em Journal of Foo}, 14(1):234--778, 2004.

\bibitem{Authors14}
Authors.
\newblock The frobnicatable foo filter, 2014.
\newblock Face and Gesture submission ID 324. Supplied as additional material
  {\tt fg324.pdf}.

\bibitem{Authors14b}
Authors.
\newblock Frobnication tutorial, 2014.
\newblock Supplied as additional material {\tt tr.pdf}.

\end{thebibliography}
}

\end{document}